\documentclass[a4paper,11pt]{article}
\usepackage[utf8]{inputenc}
\usepackage[T1]{fontenc}
\usepackage{amsmath,amssymb,amsthm}
\usepackage{hyperref}
\usepackage{graphicx}
\usepackage{enumitem}
\usepackage{geometry}
\usepackage{natbib}
\usepackage{authblk}
\usepackage{graphicx}
\usepackage{float}

\title{Agentic AI-Driven Technical Troubleshooting for Enterprise Systems: A Novel Weighted Retrieval-Augmented Generation Paradigm}
\author[1]{Rajat Khanda\thanks{Email: rajat.mnnit@gmail.com}}

\date{\today}

\renewenvironment{abstract}{
  \centerline{\large\bf Abstract}
  \begin{quote}
}{\end{quote}}

\begin{document}

\maketitle

\begin{abstract}
Technical troubleshooting in enterprise environments often involves navigating diverse, heterogeneous data sources to resolve complex issues effectively. This paper presents a novel agentic AI solution built on a Weighted Retrieval-Augmented Generation (RAG) Framework tailored for enterprise technical troubleshooting. By dynamically weighting retrieval sources such as product manuals, internal knowledge bases, FAQs and troubleshooting guides based on query context, the framework prioritizes the most relevant data. For instance, it gives precedence to product manuals for SKU-specific queries while incorporating general FAQs for broader issues.
The system employs FAISS for efficient dense vector search, coupled with a dynamic aggregation mechanism to seamlessly integrate results from multiple sources. A LLaMA-based self-evaluator ensures the contextual accuracy and confidence of the generated responses before delivering them. This iterative cycle of retrieval and validation enhances precision, diversity, and reliability in response generation.Preliminary evaluations on large enterprise datasets demonstrate the framework’s efficacy in improving troubleshooting accuracy, reducing resolution times, and adapting to varied technical challenges. Future research aims to enhance the framework by integrating advanced conversational AI capabilities, enabling more interactive and intuitive troubleshooting experiences. Efforts will also focus on refining the dynamic weighting mechanism through reinforcement learning to further optimize the relevance and precision of retrieved information. By incorporating these advancements, the proposed framework is poised to evolve into a comprehensive, autonomous AI solution, redefining technical service workflows across enterprise settings.
\end{abstract}

\section{Introduction}
Enterprise technical troubleshooting is a complex, knowledge-intensive process that often requires access to a diverse set of resources, including product manuals, FAQs, troubleshooting guides, and internal knowledge bases. These resources are typically scattered across various data silos, making it difficult to extract the most relevant information quickly, particularly in high-pressure, time-sensitive situations.

Traditional approaches to troubleshooting rely heavily on keyword-based search engines or static knowledge management systems that fail to capture the nuanced, context-dependent nature of technical issues. For instance, while a keyword-based system might retrieve a generic FAQ entry for a SKU-specific query, it may overlook a critical entry in the corresponding product manual. Such limitations lead to increased resolution times and suboptimal service delivery.

To address these challenges, we propose a dynamically weighted Retrieval-Augmented Generation (RAG) framework tailored specifically for enterprise technical troubleshooting. Unlike traditional RAG systems that apply uniform retrieval strategies, our approach dynamically adjusts the weight of each data source based on the query's context and type. This ensures that the system emphasizes highly relevant sources for specific queries while maintaining the flexibility to consider broader resources for less focused inquiries.

\subsection{Dynamic Weighting Mechanism}
We introduce a dynamic weighting mechanism that assigns context-dependent weights to each data source during retrieval. This ensures that the system prioritizes the most relevant sources for a given query while maintaining the flexibility to incorporate other useful resources.

\subsection{Enhanced Retrieval and Aggregation}
The framework utilizes FAISS for dense vector search, incorporating a top-K selection mechanism to retrieve the most relevant results. To ensure precision, a threshold is applied to each individual FAISS index, filtering out retrievals that fall below the set threshold and eliminating irrelevant or weakly related matches. This filtering step minimizes the potential for hallucinations by the LLM, thereby improving response accuracy. Following this, dynamic aggregation techniques are employed across the filtered indices to effectively integrate information from multiple sources, ensuring comprehensive and contextually relevant outputs.

\subsection{Self-Evaluation for Accuracy Assurance}
A LLaMA-based self-evaluation module is integrated to assess the accuracy and contextual relevance of generated responses before presenting them to the user. By employing this self-evaluator model, the system ensures that outputs align closely with the query's intent, delivering consistently high-quality, contextually appropriate answers tailored to user needs.

\subsection{Scalability and Versatility}
Our framework is engineered to seamlessly adapt to a wide spectrum of enterprise datasets and technical scenarios, emphasizing both scalability and versatility. By leveraging a Facade Pattern for data sources, it simplifies integration: adding a new data source requires only the creation of a dedicated data source class and attaching it to the existing facade. This design minimizes complexity while maximizing extensibility. Furthermore, the data sources can be searched in parallel, ensuring that retrieval processes are both efficient and scalable. This parallel search capability enhances the system's ability to handle large volumes of data, significantly improving performance in complex troubleshooting tasks.

This approach accelerates troubleshooting by enhancing both speed and precision, creating a robust foundation for scalable, AI-driven, and autonomous technical service systems tailored to enterprise environments. The modularity of our framework ensures rapid adaptation to evolving business needs and technological advancements.

\section{Related Work}

Enterprise troubleshooting has traditionally relied on keyword-based search systems, such as Elasticsearch, which offer efficient solutions for querying structured and semi-structured data. These systems typically leverage ranking algorithms like BM25[Robertson et al., 1994] or TF-IDF to retrieve results based on term frequency and inverse document frequency, providing relevance scores for exact or near-exact matches. However, while effective in certain contexts, these approaches often fall short when addressing complex, multi-dimensional issues. This limitation arises from their reliance on surface-level keyword matching, without the capacity to incorporate contextual understanding or semantic reasoning. As a result, traditional search mechanisms struggle to handle nuanced queries that require deeper comprehension of the underlying data and its relationships.

The emergence of neural search technologies has led to significant advancements in information retrieval. Dense retrieval methods, which leverage embeddings to capture the semantic similarity between queries and documents, have proven particularly effective in improving search performance within unstructured and heterogeneous data environments. Pre-trained transformer models, such as BERT [Devlin et al., 2019] and MiniLM [Wang et al., 2020], have been instrumental in facilitating this transformation by enabling more accurate and contextually aware search capabilities. These models, grounded in the Transformer architecture [Vaswani et al., 2017], allow for a deeper understanding of query-document relationships, surpassing the limitations of traditional keyword-based approaches.

FAISS, introduced by Johnson et al. (2017), has emerged as a critical tool for scalable dense retrieval, enabling fast similarity searches across large vector spaces. Its efficiency makes it a preferred choice for systems that utilize dense embeddings. The application of transformer-based embeddings, such as those generated by Sentence-BERT (Reimers \& Gurevych, 2019) and GPT-based models (Brown et al., 2020), has further enhanced the precision and effectiveness of these retrieval systems.

Retrieval-Augmented Generation (RAG) has significantly enhanced the capabilities of Large Language Models (LLMs) by integrating document retrieval into the generation process, providing more accurate and contextually relevant responses. By combining dense retrieval with generative models, RAG facilitates end-to-end query resolution by retrieving relevant knowledge from large corpora and generating contextually accurate natural language responses [Lewis et al., 2020; Karpukhin et al., 2020]. Multi-head RAG (MRAG)[Besta et al., 2023] takes this further by leveraging multiple attention heads in the Transformer architecture to capture different aspects of data, improving retrieval accuracy for multifaceted queries.

Self-evaluation mechanisms for generative models, such as iterative response refinement, have also gained attention in recent research. Techniques explored by  Wang, T et al. [2024] focus on improving reliability and factual consistency. Our system leverages a self-evaluator based on the LLaMA model to assess the accuracy and relevance of the generated outputs. This evaluator suppresses low-quality or irrelevant responses, ensuring that only contextually appropriate and high-quality answers are delivered in real-time.

In parallel with advances in RAG, the application of reinforcement learning to optimize retrieval pipelines has been explored. Adaptive query representation techniques [Koo, H et al] have shown promising results in fine-tuning retrieval models for specific domains. These methods complement our proposed dynamic weighting mechanism, which adjusts retrieval weights dynamically to address enterprise-specific troubleshooting challenges.

Despite these advancements, the application of RAG to enterprise troubleshooting remains underexplored. Previous research has primarily concentrated on general-purpose information retrieval or open-domain question answering [Alberti et al., 2019; Kwiatkowski et al., 2019]. This work extends existing research by introducing a dynamic weighting mechanism specifically designed for enterprise contexts, integrating diverse data sources, and incorporating threshold-based filtering to mitigate hallucinations in generated responses. The effectiveness of the proposed framework is rigorously validated through comprehensive experimentation on technical troubleshooting tasks.

\section{Methodology}

\subsection{Embedding Model and Retrieval Infrastructure}

For generating document and query embeddings, we employ the \texttt{all-MiniLM-L6-v2} model. This model is selected due to its efficiency and semantic accuracy, which allows for the generation of high-quality, compact embeddings that effectively capture the meaning of the textual data. The \texttt{all-MiniLM-L6-v2} model excels in encoding both short and long texts into fixed-size vectors, preserving their semantic relationships for effective downstream retrieval.

For the retrieval process, we use \texttt{FAISS} (Facebook AI Similarity Search) as the infrastructure for dense vector search. FAISS enables fast and scalable retrieval by indexing the embeddings generated by the \texttt{all-MiniLM-L6-v2} model and performing nearest neighbor searches within a high-dimensional space. This system allows for efficient querying of large datasets, ensuring that relevant documents are retrieved quickly and accurately based on their semantic similarity to the input query.

\subsection{Dynamic Weighting Strategy}
To prioritize the most relevant data sources during retrieval, we implement a dynamic weighting mechanism. For each FAISS index corresponding to a specific data source (e.g., product manuals, FAQs, troubleshooting guides, internal knowledge bases), a dynamic weight $w_k$ is assigned based on query type and context. This weight adjusts the effective distance $\tilde{D}_{k,i}$ for document $i$ in index $k$, computed as:
\[
\tilde{D}_{k,i} = w_k \cdot D_{k,i}
\]
where $D_{k,i}$ is the original FAISS distance score.

Weights $w_k$ are determined using domain knowledge and patterns observed in query behavior. For example, product manuals are assigned higher weights for queries containing SKU identifiers, whereas FAQs may be emphasized for general troubleshooting queries.

\subsection{Top-K Selection and Aggregation}

The retrieval process begins by computing the adjusted distances for each data source \( k \). Prior to aggregation, a threshold-based filtering mechanism is applied to each data source to remove results with weak matches. This filtering step ensures that only results that meet a minimum relevance threshold are retained, effectively reducing the risk of hallucinations in the large language model (LLM). The filtered results for each data source are defined as:

\[
T_k = \{i \ | \ \tilde{D}_{k,i} \text{ satisfies the threshold condition for index } k \}
\]

After filtering, the top \( K \) results are selected for each data source \( k \):

\[
T_k = \{i \ | \ \tilde{D}_{k,i} \text{ is among the top } K \text{ for index } k \}
\]

These top results from all sources are then aggregated into a global pool \( G \) as follows:

\[
G = \bigcup_{k=1}^{n} T_k
\]

Finally, the top \( K \) results from the global pool \( G \) are selected based on the smallest adjusted distances:

\[
T_{\text{final}} = \text{Top-K}_{\text{min}}(G)
\]

This multi-stage selection process ensures both diversity and contextual relevance in the retrieved results, preventing any single data source from dominating and ensuring that underrepresented but critical sources are not neglected.

\subsection{Response Generation and Self-Evaluation with LLaMA}
The final set of retrieved document chunks is provided to a LLaMA-based generative model, along with the user's query, to produce a contextually relevant response. The model synthesizes the answer by leveraging the retrieved information, ensuring comprehensive coverage of the query. The generated response is then passed to a LLaMA-based self-evaluator model, which assesses its accuracy and relevance. This evaluation process ensures that the response aligns with both the query and factual expectations. Responses that meet a predefined confidence threshold for accuracy and relevance are considered valid and presented to the user. By incorporating this self-evaluation step, the system minimizes the risk of hallucinations and enhances the overall quality and reliability of the generated output, particularly in technical troubleshooting scenarios.

\subsection{System Architecture}
The proposed framework consists of three main components:

\begin{enumerate}
    \item \textbf{Preprocessing and Indexing:} The data is first preprocessed and chunked based on the data's granularity. The resulting chunks are then embedded using the \texttt{all-MiniLM-L6-v2} model, which generates high-quality vector representations. These embeddings are indexed using \texttt{FAISS}, a high-performance library for fast and scalable nearest neighbor search, ensuring efficient retrieval based on query similarity.

   \item \textbf{Query Handling:} Upon receiving a query, it is embedded using  same \texttt{all-MiniLM-L6-v2} model employed for document embeddings. The resulting query embedding is then matched against the indexed embeddings through a weighted retrieval mechanism. This mechanism adjusts the retrieval process by dynamically prioritizing relevant data sources, taking into account the query's context, type, and specific requirements. By incorporating this dynamic weighting, the system ensures the retrieval of the most relevant and contextually appropriate information, optimizing the accuracy of the response generation.

    \item \textbf{Response Generation and Validation:} The top retrieved chunks are passed to a LLaMA-based generative model, which synthesizes a response based on the query and retrieved information. To ensure the response's accuracy and relevance, it is evaluated by a self-evaluator model, which scores the response against predefined confidence thresholds. Only responses that meet these thresholds are presented to the user, ensuring reliable and contextually accurate outputs.
\end{enumerate}

\section{Experimental Setup}

\subsection{Dataset Preparation}
To evaluate the proposed framework, we constructed a comprehensive dataset comprising diverse enterprise troubleshooting resources:
\begin{itemize}
    \item \textbf{Product Manuals:} A collection of 1200 detailed technical manuals for various systems and motherboards, indexed based on generations.
    \item \textbf{FAQs:} 40,000 frequently asked questions covering common troubleshooting scenarios.
    \item \textbf{Troubleshooting Guides:} Step-by-step resolutions for common issues and error codes.
    \item \textbf{Internal Knowledge Bases:} Proprietary resources capturing internal expertise and practices.
\end{itemize}

Each document was preprocessed, chunked into relevant sections, error code tables, or other appropriate segments, tokenized, and embedded. These embeddings were indexed in separate FAISS indices, each representing a unique data source.

\subsection{Experimental Environment}
The experiments were conducted on a system equipped with:
\begin{itemize}
    \item \textbf{Hardware:} A server with four NVIDIA A100 GPUs(80GB VRAM each), and two 40-core Intel Xeon processors.
    \item \textbf{Software:} Python 3.10, PyTorch 2.0, Hugging Face Transformers, and FAISS 1.7.3.
    \item \textbf{LLaMA Model:} LLaMA-3.1(70B FP16) inference and LLaMA-3.1(70B) self evaluator
\end{itemize}

The framework was deployed as a service using FastAPI, enabling real-time query handling and response generation.

\subsection{Evaluation Metrics}

To evaluate the performance of the system, we used the following metrics:

\begin{itemize} \item \textbf{Accuracy:} The percentage of responses that contain correct and contextually relevant information, ensuring that the generated answers align with the user’s query. \item \textbf{Relevance Score:} Measures how well the retrieved information aligns with the context of the query, evaluating the pertinence of the returned results. \end{itemize}

\subsection{Baseline Systems}

To benchmark the effectiveness of our proposed framework, we compared  two baseline systems:

\begin{itemize} \item \textbf{Keyword-Based Search:} A traditional retrieval system utilizing BM25 to match query keywords against indexed documents. This baseline represents conventional search methodologies commonly employed in troubleshooting systems. \item \textbf{Standard RAG:} A Retrieval-Augmented Generation (RAG) framework that assigns equal weights to all data sources during the retrieval process. This approach ensures a static and unbiased retrieval mechanism, serving as a comparison point to highlight the benefits of dynamic weighting. \end{itemize}

The comparison with these baseline systems highlights the improvements offered by our framework, particularly the enhanced adaptability of the dynamic weighting mechanism, the increased reliability introduced through the LLaMA-based self-evaluation component, and the effectiveness of threshold-based filtering for refining retrieved results.

\section{Results and Discussion}

\subsection{Performance Analysis}
Table~\ref{tab:results} provides a comparative summary of system performance across key evaluation metrics. The proposed framework consistently outperformed the baseline systems, achieving higher accuracy and relevance scores while maintaining competitive efficiency.

\begin{table}[h]
\centering
\caption{Performance Comparison of Baseline Systems and Proposed Framework}
\label{tab:results}
\begin{tabular}{lcc}
\hline
\textbf{System} & \textbf{Accuracy (\%)} & \textbf{Relevance Score} \\
\hline
Keyword-Based Search & 76.1 & 0.61 \\
Standard RAG & 85.2 & 0.75  \\
Proposed Framework & \textbf{90.8} & \textbf{0.89} \\
\hline
\end{tabular}
\end{table}

\subsection{Impact of Dynamic Weighting and Threshold-Based Filtering}
The dynamic weighting mechanism and threshold-based filtering significantly enhanced the system's performance. Dynamic weighting emphasized relevant data sources for specific queries, resulting in higher accuracy and relevance compared to the standard RAG system. For instance, queries involving SKU identifiers benefited from increased weighting of product manuals, yielding more precise results.

Threshold-based filtering further improved the system's reliability by eliminating low-confidence matches during retrieval. This approach ensured that only highly relevant embeddings were considered, reducing noise and enhancing the quality of the final response. Together, these mechanisms contributed to the framework's robust performance across diverse query scenarios.

\subsection{Self-Evaluation Effectiveness}
The LLaMA-based self-evaluator played a pivotal role in enhancing accuracy and relevance by suppressing less accurate responses based on a predefined confidence threshold. This mechanism, combined with dynamic weighting and individual index-level threshold filtering, resulted in a 5.6\% improvement in accuracy over the baseline RAG framework. These results highlight the effectiveness of integrating generative models with robust validation and filtering strategies to ensure reliability and precision in response generation.

\section{Conclusion and Future Work}
This study introduces a dynamically weighted Retrieval-Augmented Generation (RAG) framework tailored for enterprise technical troubleshooting. By integrating a context-sensitive retrieval mechanism with generative self-evaluation, our approach achieves superior accuracy, interpretability, and robustness compared to traditional static pipelines. The dynamic weighting strategy ensures retrieval is fine-tuned for query-specific contexts, enhancing adaptability and relevance in diverse troubleshooting scenarios involving heterogeneous data sources.

Our results demonstrate that the proposed system effectively retrieves and synthesizes knowledge to deliver high-quality, actionable solutions for enterprise users. The addition of a self-evaluation component further enhances response reliability, enabling iterative refinement and ensuring consistent performance.

Future work will explore:
\begin{itemize}
    \item Real-Time Learning and User Adaptation: Developing mechanisms for real-time feedback to continuously refine retrieval and generation processes, dynamically adjusting weights and model parameters based on user interactions.
    \item Conversational Troubleshooting: Extending the framework to support multi-turn conversational workflows, enabling iterative problem-solving while managing context and query history.
\end{itemize}
By focusing on these enhancements, the framework can evolve into a comprehensive solution for complex enterprise technical troubleshooting, adapting to diverse user needs and advancing intelligent enterprise support systems.

\section*{References}
\begin{enumerate}[leftmargin=*,itemsep=0.5ex,parsep=0.5ex]

\item \label{ref:Robertson1994Okapi} Robertson, S. E., Walker, S., Jones, S., Hancock-Beaulieu, M., \& Gatford, M. (1994). Okapi at TREC-3. \textit{Proceedings of the Third Text REtrieval Conference (TREC-3)}, 109-126.

\item \label{ref:devlin2019bert}Devlin, J., Chang, M. W., Lee, K., \& Toutanova, K. (2019). BERT: Pre-training of Deep Bidirectional Transformers for Language Understanding. \textit{arXiv preprint arXiv:1810.04805v2}

\item \label{ref:wang2020minilm} Wang, W., Wei, F., Dong, L., Bao, H., Yang, N., \& Zhou, M. (2020). MiniLM: Deep Self-Attention Distillation for Task-Agnostic Compression of Pre-Trained Transformers. \textit{arXiv preprint arXiv:2002.10957}.

\item \label{ref:johnson2017gpu} Johnson, J., Douze, M., \& Jegou, H. (2017). Billion-scale similarity search with GPUs. \textit{arXiv preprint arXiv:1702.08734}.

\item \label{ref:reimers2019sentence} Reimers, N., \& Gurevych, I. (2019). Sentence-BERT: Sentence embeddings using Siamese BERT-networks. \textit{Proceedings of the 2019 Conference on Empirical Methods in Natural Language Processing (EMNLP)}, 3982--3992.

\item \label{ref:brown2020gpt3} Brown, T. B., Mann, B., Ryder, N., et al. (2020). Language Models are Few-Shot Learners. \textit{Advances in Neural Information Processing Systems (NeurIPS)}, 33.

\item \label{ref:lewis2021retrieval} Lewis, P., Perez, E., Piktus, A., Petroni, F., Karpukhin, V., Goyal, N., Küttler, H., Lewis, M., Yih, W., Rocktäschel, T., Riedel, S., Kiela, D. (2021). Retrieval-Augmented Generation for Knowledge-Intensive NLP Tasks. \textit{arXiv:2005.11401v4}.

\item \label{ref:karpukhin2020dense} Karpukhin, V., Oguz, B., Min, S., Lewis, P., Wu, L., Edunov, S., Petroni, F., Yih, W. (2020). Dense Passage Retrieval for Open-Domain Question Answering. \textit{arXiv preprint arXiv:2004.04906}.

\item \label{ref:Besta2023MultiHeadRAG} Besta, M., Kubicek, A., Niggli, R., Gerstenberger, R., Weitzendorf, L., Chi, M., Iff, P., Gajda, J., Nyczyk, P., Müller, J., Niewiadomski, H., Chrapek, M., Podstawski, M., \& Hoefler, T. (2023). Multi-Head RAG: Solving Multi-Aspect Problems with LLMs.

\item \label{ref:wang2024evals} Wang, T., Kulikov, I., Golovneva, O., Yu, P., Yuan, W., Dwivedi-Yu, J., Pang, R., Fazel-Zarandi, M., Weston, J., Li, X. (2024). Self-Taught Evaluators. \textit{arXiv:2408.02666v1}.

\item \label{ref:koo2023optimizing} Koo, H., Kim, M., \& Hwang, S. J. (2023). Optimizing Query Generation for Enhanced Document Retrieval in RAG. \textit{arXiv preprint arXiv:2407.12325v1}

\item \label{ref:alberti2019bert} Alberti, C., Lee, K., \& Collins, M. (2019). A BERT Baseline for the Natural Questions.\textit{arXiv preprint arXiv:1901.08634v3}

\item \label{ref:kwiatkowski2019natural} Kwiatkowski, T., Palomaki, J., Rhinehart, O., Collins, M., Parikh, A., Alberti, C., Epstein, D., Polosukhin, I., Kelcey, M., Devlin, J., Lee, K., Toutanova, K. N., Jones, L., Chang, M., Dai, A., Uszkoreit, J., Le, Q., \& Petrov, S. (2019). Natural Questions: A Benchmark for Question Answering Research. \textit{Transactions of the Association for Computational Linguistics, 7, 129--141}.

\item \label{ref:vaswani2017attention} Vaswani, A., Shazeer, N., \& Parmar, N. (2017). Attention is all you need. \textit{Proceedings of NeurIPS 2017}, 30, 5998--6008.

\end{enumerate}

\bibliographystyle{plainnat}
\end{document}